\def\year{2021}\relax
\documentclass[letterpaper]{article} 
\usepackage{aaai21}  
\usepackage{times}  
\usepackage{helvet} 
\usepackage{courier}  
\usepackage[hyphens]{url}  
\usepackage{graphicx} 
\usepackage{natbib}  
\usepackage{caption} 
\usepackage[dvipsnames]{xcolor}
\usepackage{graphicx,subfigure}

\usepackage{amsmath, amssymb,bm}
\usepackage{hyperref}
\usepackage{enumerate,mathtools}
\usepackage{amsthm}
\usepackage{cite}
\usepackage{microtype}
\usepackage{color}
\usepackage{mathrsfs} 
\usepackage{caption}
\usepackage{xspace}
\usepackage{algorithm}      
\usepackage{algorithmic}
\usepackage{multirow}

\newcommand{\netname}{GEMNet\xspace}
\newcommand{\optmethod}{bootstrap optimization\xspace}
\newcommand{\netfullname}{Neural Architecture Search via Graph Embedding Method\xspace}
\newcommand{\workname}{NASGEM\xspace}

\newcommand{\reals}{\mathbb{R}}
\newcommand{\bEx}{\ensuremath{\mathbb{E}}}

\let\originalleft\left
\let\originalright\right
\renewcommand{\left}{\mathopen{}\mathclose\bgroup\originalleft}
\renewcommand{\right}{\aftergroup\egroup\originalright}

\DeclareMathOperator*{\argmax}{arg\,max}

\theoremstyle{definition}

\title{NASGEM: Neural Architecture Search via Graph Embedding Method}

\author {
        Hsin-Pai Cheng,\textsuperscript{\rm 1}
        Tunhou Zhang,\textsuperscript{\rm 1}
        Yixing Zhang,\textsuperscript{\rm 1}
        Shiyu Li, \textsuperscript{\rm 1} 
        Feng Liang,\textsuperscript{\rm 3} 
        Feng Yan,\textsuperscript{\rm 4} 
        Meng Li,\textsuperscript{\rm 2}
        Vikas Chandra,\textsuperscript{\rm 2}
        Hai Li,\textsuperscript{\rm 1}
        Yiran Chen,\textsuperscript{\rm 1}
        \\
}
\affiliations {
    \textsuperscript{\rm 1} Duke University \\
    \textsuperscript{\rm 2} Facebook, Inc \\
    \textsuperscript{\rm 3} Tsinghua University \\
    \textsuperscript{\rm 4} University of Nevada, Reno \\
    \{dave.cheng, tunhou.zhang, shiyu.li, hai.li, yiran.chen\}@duke.edu,
    \{meng.li, vchandra\}@fb.com,
    liangf16@tsinghua.org.cn,
    fyan@unr.edu
}
\begin{document}

\maketitle

\begin{abstract}
Neural Architecture Search (NAS) automates and prospers the design of neural networks.
Estimator-based NAS has been proposed recently to model the relationship between architectures and their performance to enable scalable and flexible search. 
However, existing estimator-based methods encode the architecture into a latent space without considering graph similarity. 
Ignoring graph similarity in node-based search space may induce a large inconsistency between similar graphs and their distance in the continuous encoding space, leading to inaccurate encoding representation and/or reduced representation capacity that can yield sub-optimal search results.
To preserve graph correlation information in encoding, we propose \workname which stands for \netfullname. \workname is driven by a novel graph embedding method equipped with similarity measures to capture the graph topology information.
By precisely estimating the graph distance and using an auxiliary Weisfeiler-Lehman kernel to guide the encoding,  \workname can utilize additional structural information to get more accurate graph representation to improve the search efficiency. 
\netname, a set of networks discovered by \workname, consistently outperforms networks crafted by existing search methods in classification tasks, i.e., with 0.4\%-3.6\% higher accuracy while having 11\%-21\% fewer Multiply-Accumulates. 
We further transfer \netname for COCO object detection. In both one-stage and two-stage detectors, our \netname surpasses its manually-crafted and automatically-searched counterparts.


\end{abstract}





\section{Introduction}
Neural architecture search (NAS)~\cite{zoph2018learning} prospers the neural architecture design process.
With the rise of NAS, automated crafted CNN models have achieved record-breaking performance for a variety of vision applications such as image classification and object detection.


The goal of NAS is to identify good architectures within a designated search space under application and resource constraints.
Earlier NAS works mainly adopt reinforcement learning~\cite{zoph2018learning}, evolutionary algorithm~\cite{real2019regularized}, bayesian optimization~\cite{kandasamy2018neural}, and differentiable~\cite{liu2018darts,luo2018neural,liang2019darts+,chen2019progressive} methods for searching. 
However, these methods usually suffer from poor search scalability. 
Fast search methods  (e.g., differentiable-based methods~\cite{liu2018darts,luo2018neural}) usually result in sub-optimal solutions 
while reward function based search methods (e.g., RL~\cite{tan2019mnasnet}) obtain high quality solutions at the cost of high computing hours. 
There is no effective way to trade-off search cost and architecture quality using these NAS methods.

To address the above issues, estimator-based methods~\cite{baker2017accelerating,Li_2020_CVPR} are proposed to enable scalable and flexible architecture search.
Estimator-based NAS formulates a representation of architecture by mapping architectures into a latent space. 
Such representation enables modeling the relationship between architecture and accuracy using an estimator, such as a supervised predictor.
The estimator-based approach is scalable as the trade-off between search cost and quality can be controlled by budgeting the number of samples for modeling the search space. 
Estimator also allows adding additional search objectives with no additional search cost.
To formulate an effective representation, recent works use graph convolutional networks (GCN) and other graph encoding schemes~\cite{li2020neural,ning2020generic} to capture the graph topology, which is important for node-based NAS.

However, existing estimator-based methods~\cite{li2020neural,ning2020generic,wen2019neural} overlook the graph distance when mapping architectures to a latent space. This results in an inaccurate 
and/or reduced representation capacity of the projected latent space. When the graph distance in the latent space cannot appropriately reflect the graph distance in the discrete space, the found cell may not be optimal.

To address the above problem, we propose \workname (Neural Architecture Search via Graph Embedding Method)
to incorporate graph kernel equipped with a similarity measure into the estimator-based search process. \workname delicately encodes graphs into a latent space and enables to search cells with high representation capacity.
The main contributions of our work can be summarized as follows:
\textbf{1)} \workname 
constructs a graphically meaningful latent space to improve the search efficiency of estimator-based method.
\textbf{2)} \workname employs an efficiency score predictor to model the relationship between cell structures and their 
performances. 
With the pretrained graph embedding, our predictor can accurately estimate the model performance based on the representation of cell structures in a graph vector.
\textbf{3) } The exploration of optimal cell structures is further improved by \optmethod, which guarantees the feasibility of graph vectors in the latent embedding space. 

Our evaluation demonstrates that \netname outperforms models obtained by other estimator-based and node-based NAS methods on multiple vision tasks with 13\%-62\% parameter reduction and 11\%- 21\% Multiply-Accumulates (MAC) reduction. Evaluation using NASBench-101 further verifies the effectiveness of our method.

\section{Related Work}
\paragraph{Graph Embedding.}
Graph embedding~\cite{goyal2018graph,grover2016node2vec} projects graph structure into a continuous latent space.
Traditional vertex graph embedding maps each node to a low-dimensional feature vector while preserving the connection relationship between vertices.
Factorization-based methods~\cite{roweis2000nonlinear}, random-walk based methods~\cite{deepwalk}, and deep-learning based methods~\cite{wang2016structural} are popular approaches used in traditional vertex graph embedding.
However, it is challenging to apply the existing graph embedding methods to deep neural networks (DNNs) for the following two reasons:
(1) similarities among cell structures of DNNs cannot be explicitly derived from traditional graph embeddings; 
(2) sophisticated deep-learning based methods like DNGR~\cite{cao2016deep} and GCN~\cite{kipf2016semi,NIPS2016_6081,li2020neural} require complex mechanisms when training on structural data.
NASGEM addresses these issues by computing the cosine similarity of two embedded graph vectors. It also facilitates the training process during the formulation of graph embedding by employing an encoding structure.

\noindent \textbf{Estimator-based NAS.}
Estimator-based NAS~\cite{wen2019neural,ning2020generic,Li_2020_CVPR} is mainly adopted to model the entire search space by leveraging the information of observed architectures.
Estimator-based NAS is able to explore architectures within unobserved search space, which is otherwise neglected by other NAS methods.
Existing estimator-based NAS works use widely adopted embedding methods such as graph convolution networks~(GCN) and autoencoders \cite{wen2019neural,li2020neural,luo2018neural}  without spending delicate efforts to improvise neural architecture representations.
Specifically, these works do not consider graph distance and similarity measures while utilizing topological information in the search space. While applied to a node-based search space, these methods usually suffer from large isomorphic graph variance in embedding space and lead to the exploration of sub-optimal blocks.
NASGEM uses a kernel-guided graph encoder to jointly learn graph topology and graph similarity while exploring architectures in the node-based search space. As a result, NASGEM enables a more precise prediction of the neural architecture performance and exploration of higher-quality building blocks.

\section{\netfullname} \label{sec:GEMoverview}

\begin{figure}
    \centering
    \includegraphics[width=0.45\textwidth]{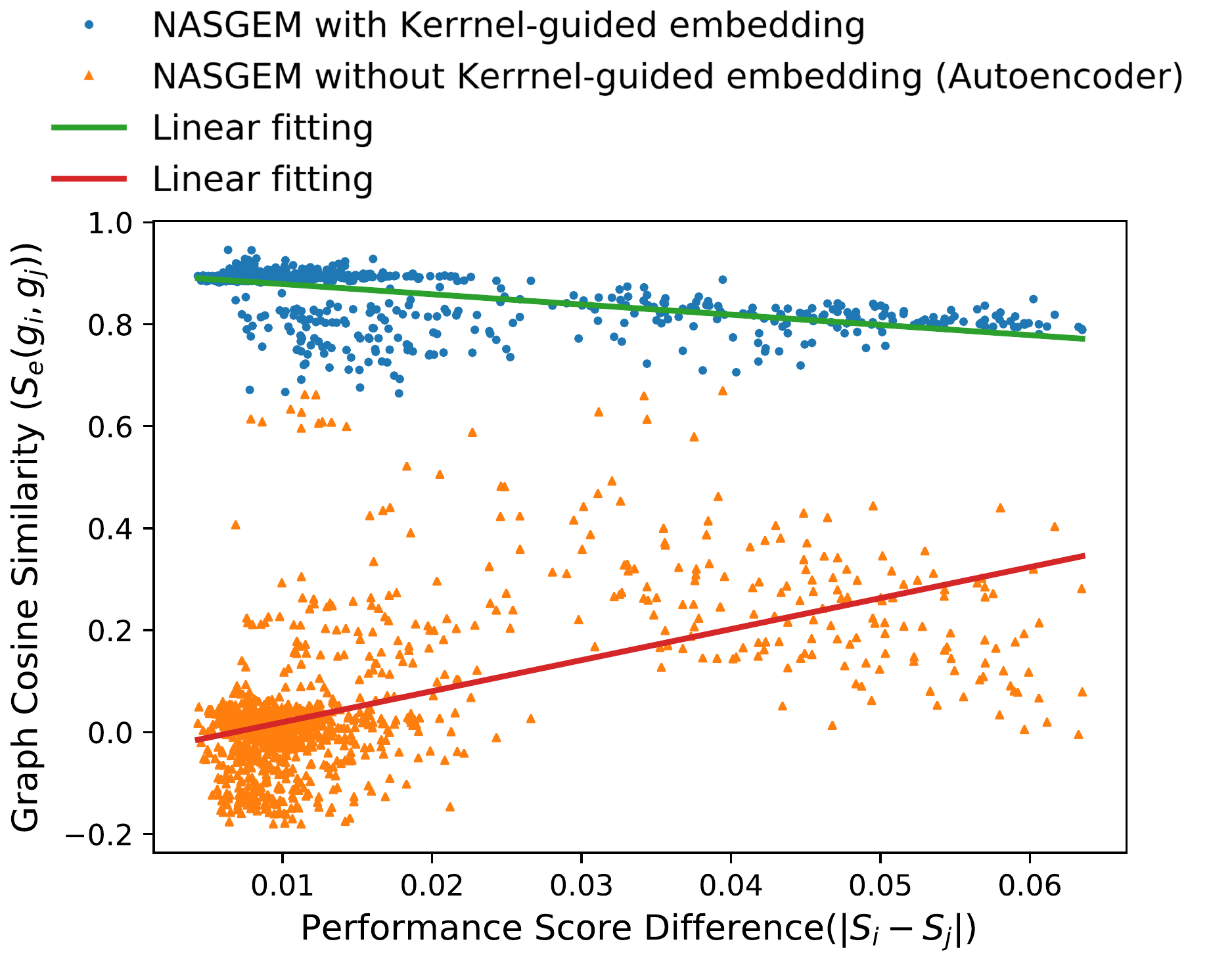}
    \caption{Each dot represents the performance score distance and the cosine similarity of vectored representation. 
    With graph embedding (blue circle dots), similar graph pairs tends to have small performance score distance.}
    \label{fig:reduce_variance}
\end{figure}

Our key intuition is that similar graphs should yield similar neural representations.
For instance, given two arbitrary graphs, the graph distance between these graphs should match their difference in the representation spaces.
However, the vanilla autoencoder~\cite{poole2014analyzing} is used by most estimator-based NAS methods~\cite{luo2018neural,zhang2019d} and overlook such topological information.
As shown in Fig.~\ref{fig:reduce_variance}, 
the aforementioned autoencoder fails to exploit the negative correlation (orange triangle) between performance score difference and pairwise graph similarity.
The incorrect correlation is due to the arbitrary representation learned by the autoencoder, therefore we envision a kernel-guided mechanism to formulate an embedding that can preserve topological information in the learned neural representation.
This motivates us to develop NASGEM, a node-based neural architecture search method composed of a kernel-guided encoder to learn an effective embedding, an estimator built upon the embedding to utilize topological information, and a bootstrap optimization approach to finalize the design of high-performing neural architectures. 

\begin{figure*}[t]
\begin{center}
   \includegraphics[width=0.99\linewidth]{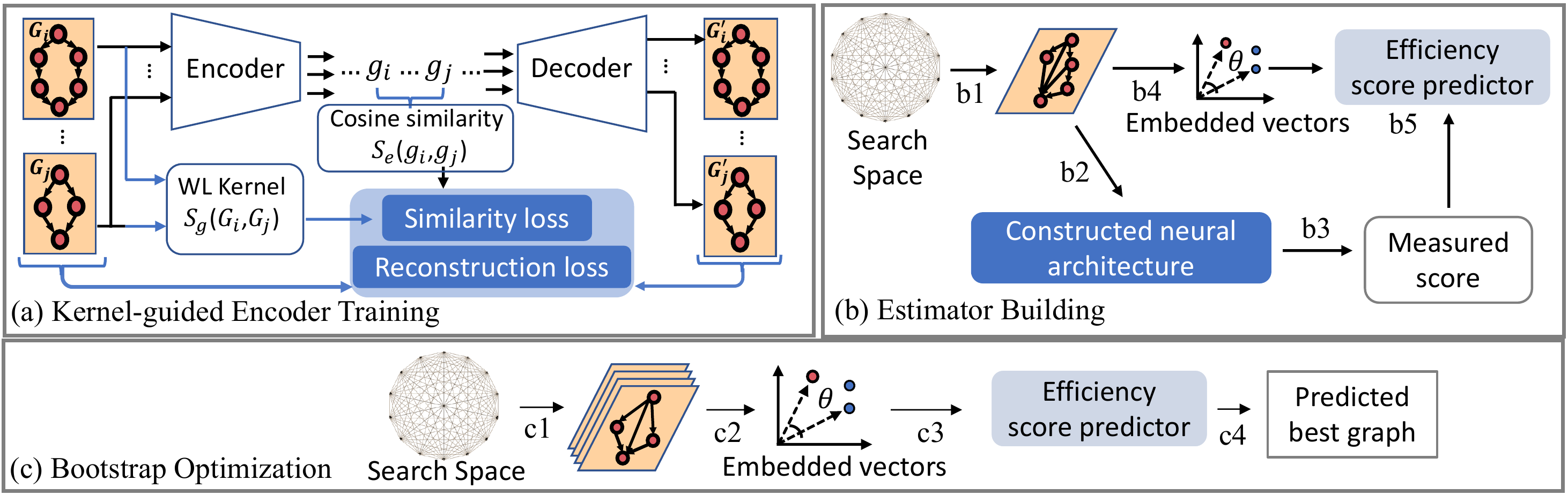}
\end{center}
\vspace{-1em}
\caption{
Workflow of \workname. (a) Encoder Training: the encoder learns to map graphs into a continuous embedding space by jointly minimizing the reconstruction loss and graph similarity loss. 
(b) Estimator Building: we \textit{(b1)} randomly sample graphs from the search space; then \textit{(b2)} build these graphs into neural networks and \textit{(b3)} measure their efficiency scores; \textit{(b4)} we also embed each sub-graph into a continuous vector with the trained kernel-guided encoder; finally, \textit{(b5)} we train the efficiency score predictor with the embedded vectors and the corresponding efficiency score. 
(c) Bootstrap Optimization: we \textit{(c1)} sample a large amount of graphs from the search space, \textit{(c2)} embed then into vectors with the kernel-guided encoder and \textit{(c3)} obtain the predicted efficiency score; \textit{(c4)} we select the graph with the highest predicted score as the candidate.
} 
\label{fig:framework}
\end{figure*}

 
Fig.~\ref{fig:framework} depicts the 3-step workflow of \workname.
In the first step, we construct a kernel-guided encoder to derive graph vectors from candidate graphs. 
The encoder is trained to jointly minimize reconstruction loss and pairwise graph similarity loss, see Figure ~\ref{fig:framework}(a).
In the second step, we utilize the pretrained graph encoder to model the relationship between graph vectors and their corresponding performance. 
An efficiency score predictor is introduced as an estimator during the exploration process, see Figure ~\ref{fig:framework}(b).
Finally, we obtain the optimal cell structure by applying \optmethod in a large sample space.
The cell structure with the highest score determined by the efficiency score predictor is adopted as the optimal building block, 
see Fig.~\ref{fig:framework}(c).


\subsection{Kernel-guided Encoder Training} \label{sec:graphembedding}
\begin{figure}{}{}
    \begin{center}
    \includegraphics[width=0.88\linewidth]{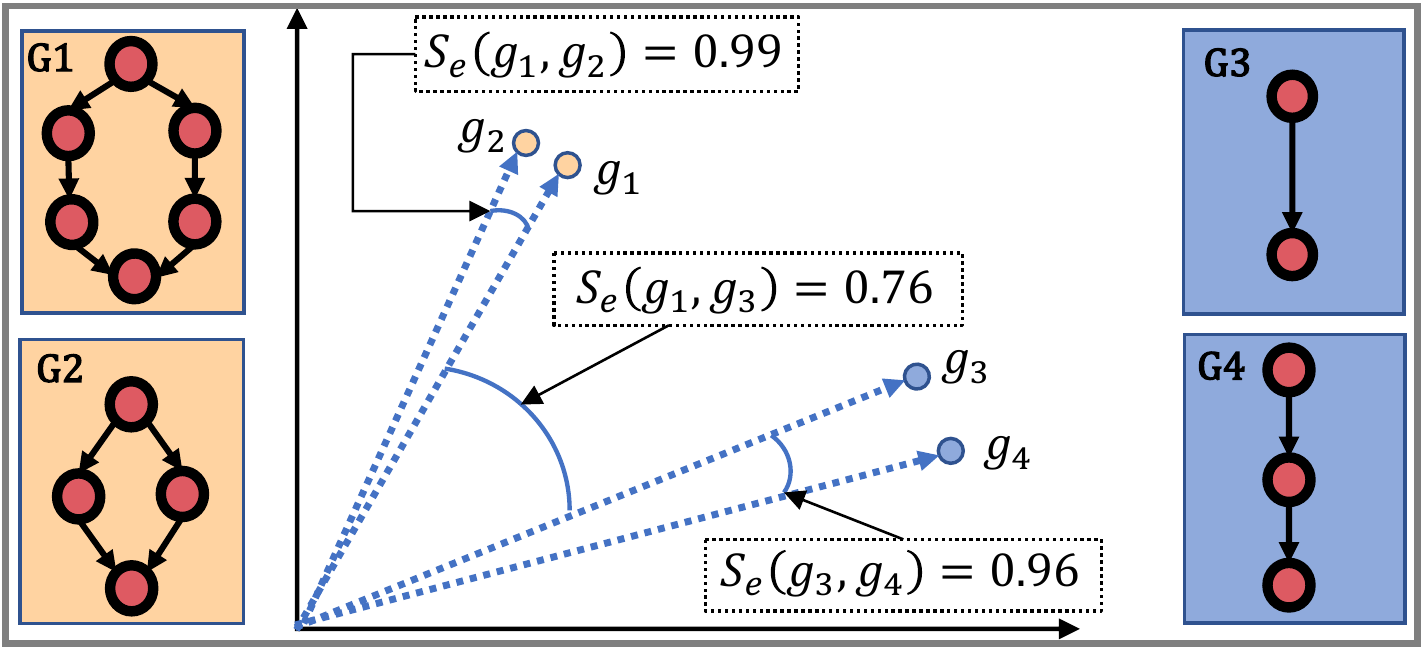}
    \end{center}
    \vspace{-1em}
        \caption{Our goal is to construct an embedding space such that the distance of graphs in the embedding space reflects its graph similarity. Graphs with more graph similarity (e.g., G3 and G4) have closer distance ($S_g$) in the embedding space.}
        
    \label{fig:embedding_vis}
\end{figure}

In step (a) of \workname workflow, we train kernel-guided encoder, $\mathbf{E}: \mathbb{R}^{n\times n}\longrightarrow\mathbb{R}^{d}$, to vectorize the adjacency matrix of a graph to an inner product space that can represent the topological graph structure, node information (DNN operation), neighboring connections (distribution of tensors), and the pairwise graph similarity. 
 



To measure the similarity of graphs in discrete topological space, there are various approaches in graph theory, such as WL kernel~\cite{shervashidze2011weisfeiler}.
For continuous graph embedding space, cosine similarity is widely used to measure the similarity of graph vectors in $[-1,1]$.
The encoded vectors aim at preserving graph similarity measured by cosine similarity in the continuous space, see Fig.~\ref{fig:embedding_vis}. Therefore, we train the encoder with the objective of minimizing the difference between WL kernel value in the discrete graph space and the cosine similarity value in the continuous space.

\noindent \textbf{Weisfeiler-Lehman (WL) Kernel.}
To estimate graph similarity, a common method is to learn a positive definite kernel, $k : \mathcal{X} \times \mathcal{X} \rightarrow \reals$, where  $\mathcal{X}$ represents the adjacency matrix set. Here we adopt Weisfeiler-Lehman (WL) kernel~\cite{shervashidze2011weisfeiler} as a graph similarity measure of two arbitrary input graphs $\mathbf{G}_i, \mathbf{G}_j$ as follows:
\begin{equation}
    S_g(\mathbf{G}_i, \mathbf{G}_j) = k_{WL}^{(h)}(\mathbf{A}_i, \mathbf{A}_j),
\end{equation}
where $\mathbf{A}_i, \mathbf{A}_j$ are the adjacency matrices for graph $\mathbf{G}_i$, $\mathbf{G}_j$. $h$ denotes the iteration times of computing WL kernel. 
For graphs with $N$ nodes, the WL kernel can be computed with $h$ iterations in $O(h \times N)$. Compared with other graph similarity metrics, WL kernel is able to measure large and complex graphs with relatively low computational complexity.
The procedure and code of measuring the similarity between two input graphs is provided in the supplementary material.

\noindent \textbf{Similarity measure of graph vectors.} 
We adopt cosine similarity to measure the similarity of the vectorized graph representations (i.e., graph vectors). 
The cosine similarity measure on the embedding space is defined as:
\begin{equation}
S_{e}( \mathbf{g}_i, \mathbf{g}_j) = \frac{ \mathbf{g}_i\cdot \mathbf{g}_j}{||\mathbf{g}_i||~ ||\mathbf{g}_j||} ,
\end{equation}
 where $\mathbf{g}_i$ and $\mathbf{g}_j$ are vectorized representation of graph $\mathbf{G}_i$ and $\mathbf{G}_j$. 
The direction of the vector reflects the proximities of the original graphs. Compared with the standard Euclidean distance in $\mathbb{R}^n$, cosine similarity is scale-invariant. In addition, since the range of cosine distance is bounded, it is comparable with the similarity value given by WL kernel. Thus we can supervise the training of the embedding function with the difference between the cosine similarity value and the WL kernel value.

\noindent \textbf{Encoder training objective.} As graph similarities are always measured pairwise, the encoder is trained and evaluated on a large number of graph pairs ($\mathbf{G}_i$, $\mathbf{G}_j$). 
To formulate a mapping to represent topological graph structure in a continuous space, the cosine similarity of the encoded graph vectors shall represent the similarity of the corresponding original graph in the original discrete topological space. 
To satisfy the above training objectives, we define the similarity loss with respect to a pair of input graph: ($\mathbf{G}_i$, $\mathbf{G}_j$) as:
\begin{equation}
\mathcal{L}_s(\mathbf{G}_i, \mathbf{G}_j;\mathbf{E}) = [S_e(\mathbf{E}(\mathbf{A}_i),\mathbf{E}(\mathbf{A}_j))-S_g(\mathbf{G}_i, \mathbf{G}_j)]^{2} .
\end{equation}

Besides similarity loss, we also consider reconstruction loss, $\mathcal{L}_r(\mathbf{G}_{i},\mathbf{D}(\mathbf{E}(\mathbf{G}_{i})))$,  for each graph. Here $\mathbf{D}$ is a decoder that maps the encoded graph to the same dimension as $\mathbf{G_i}$ Therefore, our final loss function of encoder training is as follow:
\begin{equation}
\begin{split}
\mathbf{E}^{*} &= \min_{\mathbf{E}} \sum_{i,j}\left\{{\mathcal{L}_s(\mathbf{G}_i, \mathbf{G}_j;\mathbf{E})}\right.\\
&+ \left.\mathcal{L}_r(\mathbf{G}_{i},\mathbf{D}(\mathbf{E}(\mathbf{G}_{i}))) + \mathcal{L}_r(\mathbf{G}_{j},\mathbf{D}(\mathbf{E}(\mathbf{G}_{j})))\right\}.
\end{split}
\end{equation}


Following common practice in autoencoders \cite{hinton1994autoencoders}, we adopt fully-connected feedforward networks for both encoder and decoder.
The encoder is trained independently before the searching procedure. 
We randomly generate a number of graph structure pairs with the same numbers of nodes and apply WL kernel to provide ground truth labels. Details and code implementation can be found in supplementary material.

\subsection{Estimator Building}
\label{sec:cellexploration}

\noindent \textbf{Performance estimation of graph vectors.}
Like most of the previous works \cite{liu2018darts,pham2018efficient}, we map DAGs to cell structures, which are used as building blocks of DNNs. 
Each node in the DAG stands for a valid DNN operation (such as depth-wise separable convolution 3$\times$3), and each edge represents the flow of tensors from one node to another.
When mapping cell structures to DNN architectures, nodes with no input connections (i.e., zero in-degree) are dropped while nodes with in-degree larger than one is inserted a concatenate operation.
The output of building blocks can be constructed from the leaf nodes with zero out-degree.
The concatenation of these leaf nodes along the last dimension gives the output feature maps.

To evaluate the performance of DNN architectures formed by the corresponding cell structures of graph vectors, we use efficiency score instead of accuracy as our search metric so that both performance and efficiency are taken into considerations. The efficiency score for candidate graph $\mathbf{G}$ is formulated as:
\begin{equation}\label{eq:eff_score}
    S(\mathbf{G}) = ACC[N(\mathbf{G})] - \lambda\log(MAC[N(\mathbf{G})]),
\end{equation}
where $N(\mathbf{G})$ is the DNN constructed with the cell represented by $\mathbf{G}$. $ACC$ is the validation accuracy on proxy dataset. $MAC$ is the number of Multiply-Add operations of $N(\mathbf{G})$ measured in Millions. $\lambda$ is a penalty coefficient. 
We use MAC as the penalty term since it can be precisely measured across search iterations. 
Most compact models~\cite{howard2017mobilenets,sandler2018mobilenetv2,tan2019mnasnet} have hundred millions of MACs while large models~\cite{szegedy2015going} can have billions of MACs. 
This penalty function can urge the search iteration towards improving the performance of small models or deflating the complexity of large models, and thus strike a balance between complexity and performance.


\noindent \textbf{Train Efficiency Score Predictor.}
The efficiency score predictor $\textbf{P}:\mathbb{R}^d\longrightarrow\mathbb{R}$ maps the $d$-dimensional graph vector to a real-value that indicates the performance of architectures built upon this graph vector.  
The predictor is a fully connected neural network with activation function ReLU.
We maintain a set $\{(g, y)\}$, where $g$ is a graph vector and $y$ is its efficiency score measured on proxy dataset. 
In each iteration, we add current selected graph vector/score pair into the set and train the predictor $P$ with this enlarged set. 
The predictor becomes more accurate by using the efficiency score of new samples for fine-tuning.
More importantly, the predictor in \workname is built on top a smoother latent space, which is constructed through the graph embedding of unrestricted DAGs. Such accurate prediction allows to explore a wider search space with extremely small search cost (0.4 GPU days).
Predictor training algorithm and code implementation can be found in Supplementary Material.

Fig.~\ref{fig:reduce_variance} shows the feasibility of the proposed predictor. 
We can see the cosine similarity of the output vectors of graph encoder is inversely proportional to their performance distance. This indicates that after training, kernel-guided encoder can learn to encode similarity information between two graphs into the intersection angle of their vectorized representations.
In other words, graph embedding can improve the accuracy of predictor by enhancing robustness under isomorphic graphs.

We also show that for isomorphic or similar graphs, our predictor gives close performance score. 
For a fully connect neural network, its Lipschitz constant always exists~\cite{jordan2020exactly,virmaux2018lipschitz}. Thus, we assume the Lipschitz constant of predictor $\textbf{P}$ after training is $K$, which means $\forall x_1, x_2 \in \reals^d$,
\begin{align}\label{eq_Lipschitz_constant}
    \left|\textbf{P}\left(x_{1}\right) - \textbf{P}\left(x_{2}\right)\right| \leq K \left\|x_{1} - x_{2}\right\|
\end{align}
where $\| \cdot \|$ represents $L^2$ norm. Let the input of predictor be a random vector $X\sim \mu \in\reals^d$. $X^{\prime}$ and $X$ are independently drawn from probability measure $\mu$. We have 
\begin{align}\label{eq_step_1_b}
   &\bEx_{\mu\times \mu}\left[ \left|\textbf{P}\left(X\right) - \textbf{P}\left(X^{\prime}\right)\right|^2\right] \\
   &=  \bEx_{\mu\times \mu}\left[ \textbf{P}^2\left(X\right) + \textbf{P}^2\left(X^{\prime}\right) - 2\textbf{P}\left(X\right)\textbf{P}\left(X^{\prime}\right)\right]\\
   &=  2\bEx\left[\textbf{P}^2(X)\right]-2\left(\bEx\left[\textbf{P}(X)\right]\right)^2 \\
   &= 2 \bEx\left[\left(\textbf{P}(X) - \bEx\left[\textbf{P}(X)\right]\right)^2 \right]\label{eq_step_1_e}.
\end{align}
Based on \eqref{eq_Lipschitz_constant}, 
\begin{align}\label{eq_step_2_b}
   \bEx_{\mu\times \mu}\left[ \left|\textbf{P}\left(X\right) - \textbf{P}\left(X^{\prime}\right)\right|^2\right] & \le K^2  \bEx_{\mu\times \mu}\left[  \left\|X - X^{\prime}\right\|^2\right]\\
   & {\kern -50pt}= 2K^2 \bEx\left[\left(X - \bEx\left[X\right]\right)^\top\left(X - \bEx\left[X\right]\right) \right]\label{eq_step_2_e}.
\end{align}
Based on \eqref{eq_step_1_b}-\eqref{eq_step_1_e} and \eqref{eq_step_2_b}-\eqref{eq_step_2_e},
\begin{align}\label{eq:variance_preditor}
    {\kern -15pt}\bEx\left[\left(\textbf{P}(X) - \bEx\left[\textbf{P}(X)\right]\right)^2 \right] \le K^2 \bEx\left[\left\|X - \bEx\left[X\right]\right\|^2\right].
\end{align}
Equation ~\eqref{eq:variance_preditor} shows the variance of the output (performance score) of predictor is upper bounded by its Lipschitz constant and the variance of input vectors. After training, the predictor is determined with a fixed Lipschitz constant $K$. Encoder with WL kernel embedding decreases the variance of input for isomorphic or similar graphs. Therefore, it enhances the robustness of predictor under isomorphic or similar graphs.

\subsection{Bootstrap Optimization}
After the predictor is trained using a large number of neural architectures and their corresponding efficiency scores, our goal of finding the optimal cell structure in the topological graph space is equivalent to finding the graph vector that has the highest score according to the efficiency score predictor, formulated as:
\begin{equation}
    g^{*}=\argmax_{g}{\textbf{P}(g)},
\end{equation}
where $g=\mathbf{E}^{*}(\textbf{A})$ is the embedded graph vector after passing adjacency matrix \textbf{A} into the pretrained graph encoder, and $\textbf{P}(g):\mathbb{R}^d\longrightarrow\mathbb{R}$ is the efficiency score predictor that estimates the efficiency score of a given cell $A$. 

\workname explores a continuous immense search space consisting of hyper-complex families of cell structures.
For efficient exploration, we introduce an exploration method based on two empirical beliefs: (1) Optimal cell structures within the search space is not unique as various architectures of the similar isomorphism can yield equally competitive results.
(2) Finding the optimal graph vector in the continuous search space and then decoding may not discover a valid architecture, as the mapping from graph vectors to discrete topological cell structures may not be injective.

For simplicity, we use \textit{Bootstrap Optimization} to address the above issues by sampling the cell structures with replacement among a large sample space \textbf{S} and picking up the best one as our post-searching approximation. 
With the pretrained graph encoder $\mathbf{E}$, we randomly sample cell structure $A\in \textbf{S}$ from the sample space and approximate the best candidate cell structure $A^{*}$ by predicting the efficiency score using the efficiency score predictor $\textbf{P}$:
\begin{equation}
    A^{*}=\argmax_{A\in \textbf{S}}\textbf{P}(\mathbf{E}(A)).
\end{equation}

\section{Experimental Evaluation}\label{sec:expt}

\begin{figure*}[t] 
\centering     
    \subfigure[
    With graph kernel guided embedding. ]{\label{fig:a}\includegraphics[width=0.35\textwidth]{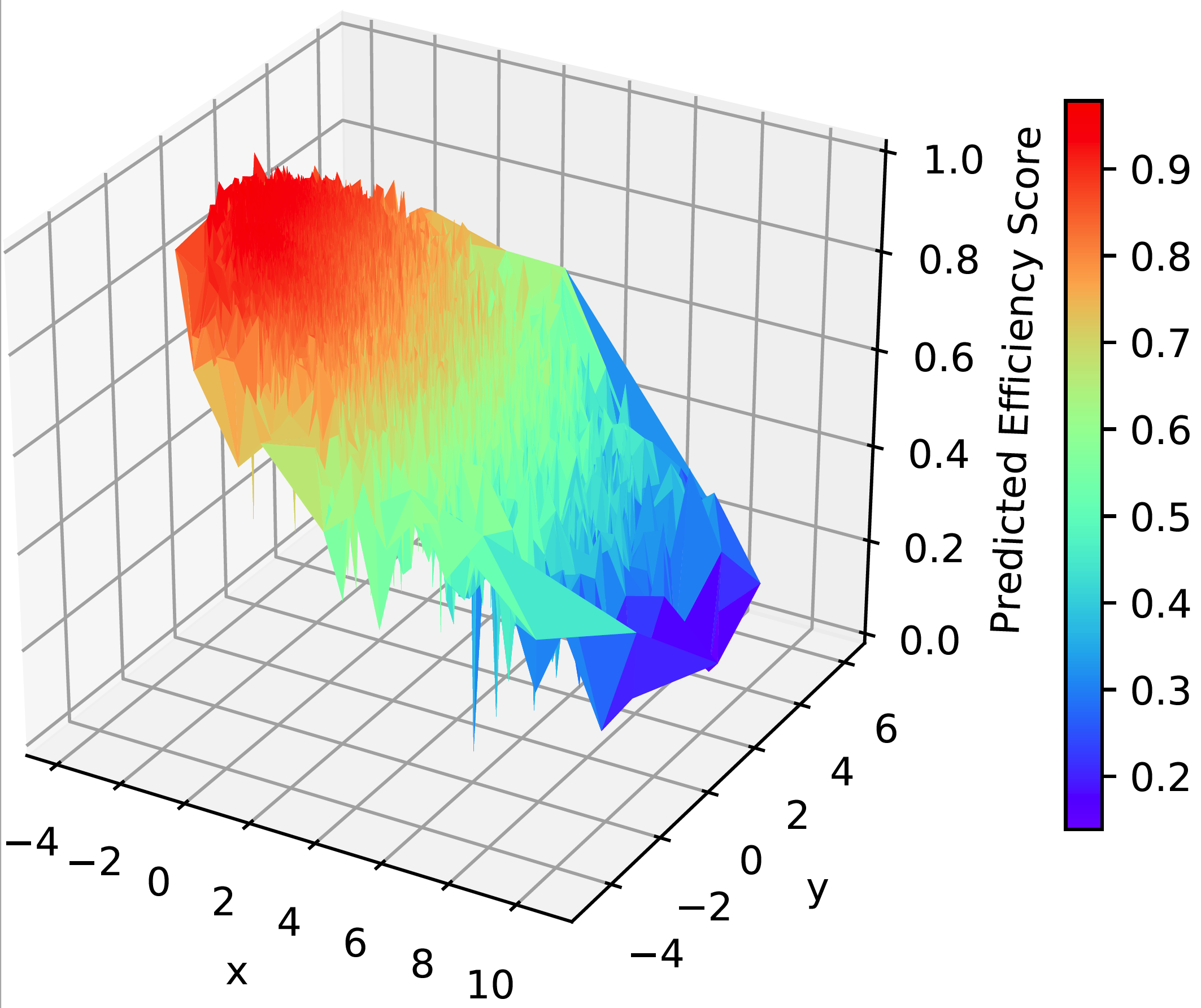}}
    \subfigure[
    Without graph kernel guided embedding. ]{\label{fig:b}\includegraphics[width=0.35  \textwidth]{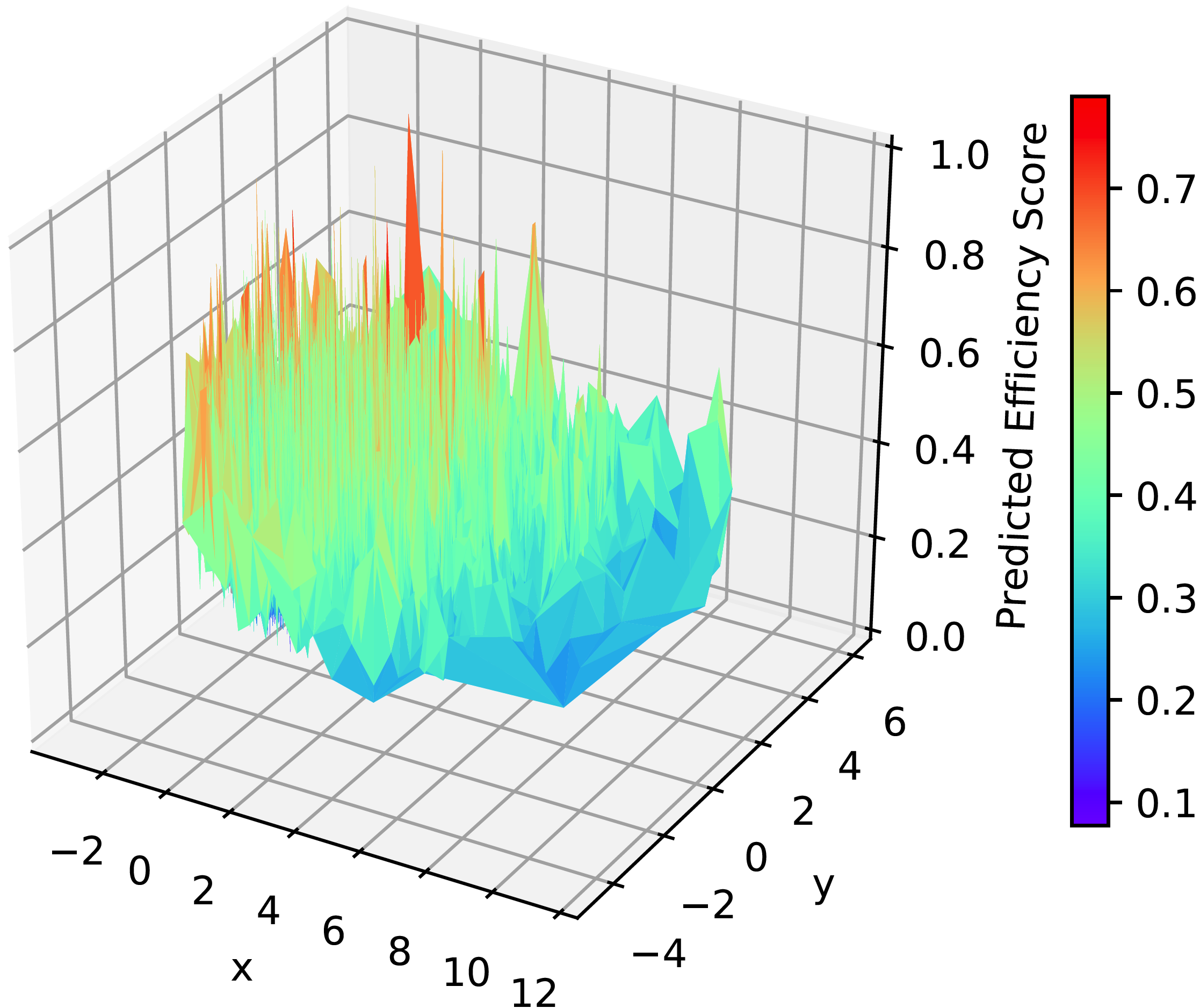}}
\caption{Efficiency score surface of efficiency score predictor based on DNN performance on CIFAR-10 dataset and computation cost in terms of MACs. We use PCA to project the graph vectors into a 2-dimensional space named x and y and plot the efficiency score of graph vector using a heatmap for (a) with graph kernel guided embedding and (b) the case of without embedding.}\label{fig:predictor_surface}
\end{figure*}


We apply \workname to search efficient mobile neural architectures 
on image classification and object detection tasks.
We initiate architecture search with a complete DAG of $N=30$ so that a rich source of cells can be constructed. 
Each node can choose from either 1$\times$1 convolution or 3$\times$3 depthwise separable convolution as its candidate operation.
Following the common practice in NAS, each convolution operation adopts a Convolution-BatchNorm-ReLU triplet~\cite{Xie_2019_ICCV,liu2018darts}.
We use $1/10$ of the CIFAR-10 dataset as a proxy to evaluate performance, and use the obtained performance to train the predictor.




\subsection{Learning Dynamics of \workname}

We illustrate the learning dynamics of \workname by plotting the efficiency score surface with respect to latent vectors in the graph embedding space.
The efficiency score surface is obtained by sampling 50,000 adjacency matrices, mapping them to the continuous embedding space, and passing them through the efficiency score predictor.

Fig.~\ref{fig:predictor_surface} illustrates the efficiency score surface of efficiency score predictors with and without using graph embedding as a latent vector. 
This can be interpreted as the embedding of the optimal cell structure in the continuous embedding space. 
Under the kernel-guided embedding, the efficiency score surface is smoother, which achieves global optimum more easily and more efficiently.
In contrast, without graph kernel guided embedding, the efficiency score surface cannot be constructed smoothly, which makes the optimization process more difficult and less efficient.





\subsection{Classification on ImageNet}\label{sec:classification_exp}
Table \ref{tab:imagenet_results} summarizes the key performance metrics on the ImageNet dataset.
For a fair comparison, we compare with the most relevant NAS works that also use either encoding scheme or node-based search space (without predefined chain-liked backbone). Note that we report the total search cost for all phases: encoder training, estimator building, and bootstrap optimization. 
\netname, the optimal architectures crafted by our explored building block, are consistently more accurate with fewer parameters and MACs. Specifically, \netname-A and \netname-B surpass all the node-based works (DARTS, P-DARTS, PC-DARTS, SNAS, GDAS, MiLeNAS). \netname with kernel-guided graph embedding also outperforms the most recent NAS methods with alternative embedding methods such as LSTM, GCN, and GATES. 

 \begin{table*}[h!]
    \begin{center}
    \scalebox{1}{
    \begin{minipage}{\textwidth}

    \begin{tabular}{c|c|cc|c|c|c}
    \hline
     \multirow{2}{*}{\textbf{Architecture}} & \textbf{Encoder}  & \multicolumn{2}{|c|}{\textbf{Test Err.(\%)}}  &  \textbf{\#Params} & \textbf{MACs}  & \textbf{Search Cost} \\
      &  &  \textbf{top-1} & \textbf{top-5} &  \textbf{(M)} & \textbf{(M)}  & \textbf{(GPU days)}  \\
    \hline\hline

            
            RandWire-WS~\cite{Xie_2019_ICCV} & / & 25.3\tiny{$\pm$0.25} & 7.8\tiny{$\pm$0.15} & 5.6\tiny{$\pm0.1$} & 583\tiny{$\pm6.2$} & / \\
            DARTS~\cite{liu2018darts} & / & 26.7 & 8.7 & 4.7 & 574 & 4.0 \\
            P-DARTS~\cite{chen2019progressive} & / & 24.7 & 7.5 &  5.1 & 577 & 0.3 \\
            PC-DARTS~\cite{xu2019pc} & / & 25.1 & 7.8 & 5.3 & 586 & 0.1 \\
            SNAS~\cite{xie2018snas} & / & 27.3 & 9.2 & 4.3 & 522 & 1.5 \\
            GDAS~\cite{dong2019searching} & / &26.0 & 8.5 & 5.3& 581& 0.21 \\
            MiLeNAS~\cite{he2020milinas} & / &25.4 & 7.9 & 4.9& 570& 0.3\\
            BayesNAS~\cite{Zhou2019BayesNASAB} & / &26.5 & 8.9 & 3.9& /& 0.2 \\
            NAONet~\cite{luo2018neural}& LSTM & 25.7 & 8.2 & 11.35 & 584 & 200\\
            NGE~\cite{li2020neural} & GCN &25.3 & 7.9 & 5.0& 563& 0.1\\
            GATES~\cite{ning2020generic} & GATES & 24.1 & / & 
            5.6& /&/\\ 
            \textbf{\netname-A (Ours)} & Kernel-guided MLP & \textbf{23.7} & \textbf{7.9} & \textbf{4.3} & \textbf{463}& ~\textbf{0.4}\\
            \textbf{\netname-B (Ours)} & Kernel-guided MLP & \textbf{23.3}  & \textbf{7.8} & \textbf{4.5} & \textbf{563}& ~\textbf{0.4}\\
            \hline

    \end{tabular}
    \end{minipage}
    }
    \end{center}
    \caption{ImageNet results with different computation budget. For fair comparison, the input image resolution is fixed at 224$\times$224. Note that the pareto frontier of MACs and parameter count (\#Params) is not linear. Further reducing computation cost under a small computation regime is very challenging as redundancy is already low. 
    }  
    \label{tab:imagenet_results}
    \vspace{-1em}
\end{table*}

\subsection{Object Detection on COCO}
To further evaluate the transferability of \netname, we conduct object detection experiments on the challenging MS COCO dataset~\cite{lin2014coco}. We use the whole COCO \emph{trainval135} as training set and validate on COCO \emph{minival}. For both two-stage Faster RCNN detector with Feature Pyramid Networks (FPN)~\cite{fasterrcnn, fpn} and one-stage RetinaNet~\cite{retinanet} detector, the input images are resized to a short side of 800 pixels and a long side not exceeding 1333 pixels.
As shown in Table \ref{tab:object_detection}, compared with the manually crafted MobileNetV2~\cite{sandler2018mobilenetv2} and automatically searched MnasNet~\cite{tan2019mnasnet}, \netname detection model achieves up to 0.7\% higher AP with fewer parameters and lower MACs. 

\begin{table*}[h]
\centering

\begin{tabular}{c||c|c|c|c|c|c|c|c|c} 
\hline
\textbf{Backbone}  & \textbf{Detector}  & \textbf{Params(M)} & \textbf{MACs(G)} & \textbf{AP$_s$}   & \textbf{AP$_m$}  & \textbf{AP$_l$}  & \textbf{AP$_{50}$}  & \textbf{AP$_{75}$}  & \textbf{AP}    \\ 
\hline \hline
MobileNetV2$\times$1.4      & \multirow{4}{*}{RetinaNet}   & 14.2      & 170.5   &  18.7 & 36.6 & 44.5 & 52.9 & 35.7 & 33.5 \\
MnasNet$\times$1.3           &                              & 14.8      & 169.2   &  19.7 & 38.2 & 47.6 & 55.2 & 36.9  & 35.0 \\
\textbf{\netname-A (Ours)}   &                              & \textbf{13.5}      & \textbf{166.9}   &  \textbf{21.2} & 38.9 & 47.3 & 55.6 & 37.5 & 35.4 \\
\textbf{\netname-B (Ours)}   &                              & 14.2      & 169.2   &  21.1 & \textbf{39.3} & \textbf{47.9} & \textbf{55.8} & \textbf{38.2} & \textbf{35.7} \\ 
\hline
MobileNetV2$\times$1.4       & \multirow{4}{*}{Faster RCNN} & 22.1      & 133.4  & 19.4 & 36.5 & 43.6 & 55.6 & 35.5 & 33.6  \\
MnasNet$\times$1.3           &                              & 22.6      & \textbf{132.1}   & 21.4 & 38.6 & 45.9 & 57.9 & 37.5 & 35.4  \\
\textbf{\netname-A (Ours)}   &                              & \textbf{21.4}      & 133.1   & 21.7 & 38.7 & 45.1 & 57.6 & 37.8 & 35.3  \\
\textbf{\netname-B (Ours)}   &                              & 22.0      & 133.6   & \textbf{21.9} & \textbf{39.0} & \textbf{45.9} & \textbf{58.2} & \textbf{38.1} & \textbf{35.7}  \\
\hline
\end{tabular}
\caption{Results on MS COCO dataset. Parameters and MACs are measured on the whole detector with input size $800\times1333$.
} 
\label{tab:object_detection}
\vspace{-1em}
\end{table*}



\subsection{Evaluation on NASBench-101}

To facilitate the reproducibility of NAS and evaluate search strategy, NASBench-101~\cite{pmlr-v97-ying19a} provides abundant results for various neural architectures (about 423K) in a large number of search spaces.
To justify the effectiveness of \workname, we further perform evaluation on NASBench-101.
We randomly sample a fixed number of candidate architectures (200$\sim$2000) from NASBench-101 within a fixed operation list given by NASBench-101.
Then we train the efficiency score predictor with/without our proposed graph embedding on these sampled architecture respectively.
Finally, we select the best architecture through bootstrap optimization by using the trained efficiency score predictor to evaluate a total of 50K architectures and pick the best one.



We measure the performance of NASGEM on NASBench-101 by \textit{Global Prediction Bias}, $B = A_B - A_P$. Here $A_B$ is the \textit{global accuracy}, which is the best accuracy in the whole search space given in NASBench-101. $A_P$ is the \textit{predicted accuracy}, which is the accuracy achieved by \workname. When \textit{predicted accuracy} is very close to \textit{global accuracy}, \textit{global prediction bias} ($B$) is moving toward zero. Therefore, the smaller $B$ reflects the more accurate prediction.

\begin{figure}[h!]{}{}
\centering
\includegraphics[width=0.6\linewidth]{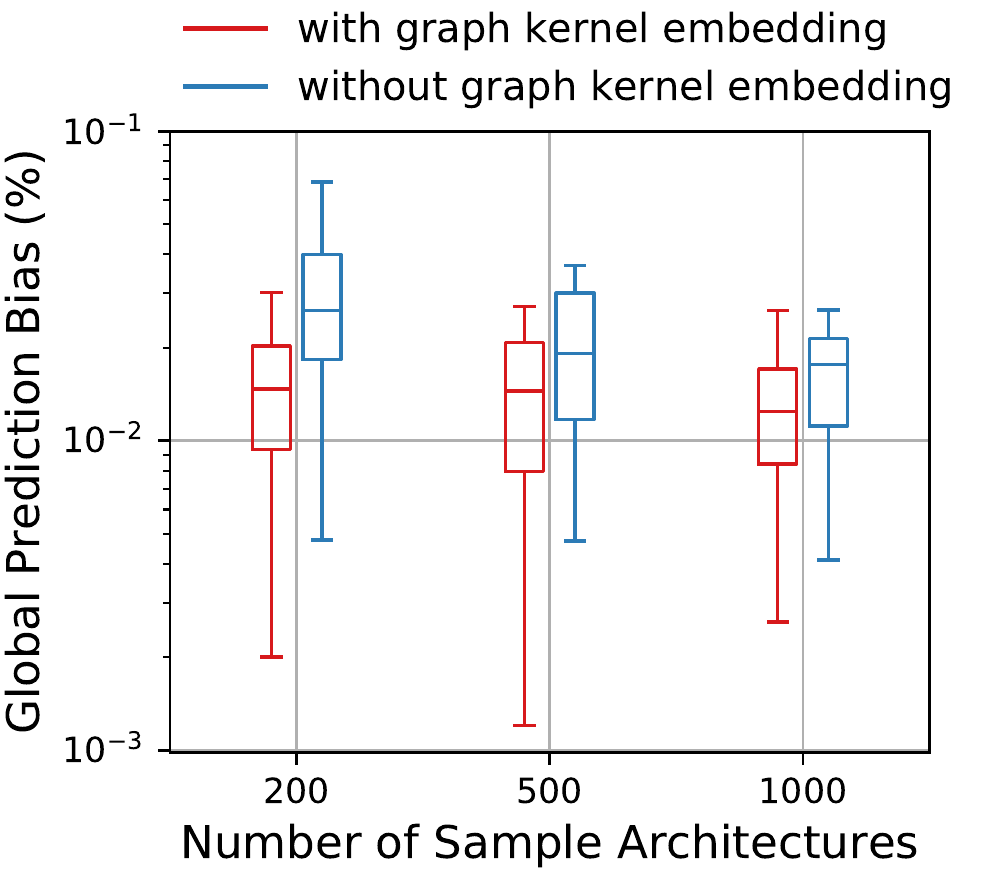}
\vspace{-3mm}
\caption{Efficiency score predictor's performance on NASBench-101 with and without graph kernel embedding.}
\label{fig:NASGEM_on_nasbench}
\vspace{-2mm}
\end{figure}

As shown in Fig.~\ref{fig:NASGEM_on_nasbench}, with the guidance of graph kernel embedding, there is around 0.2\% gain on predicting global accuracy on the NASBench-101 dataset.
Such topological information can facilitate the training of efficiency score predictor given insufficient data.
NASGEM also produces more stable results as structural knowledge represented by graph embedding generalizes better than binary adjacency matrices. Thus, the performance of NASGEM is less sensitive to the number of evaluated samples in the search space.

\subsection{Ablation Studies}

\begin{figure}[h!]{}{}
\centering
\includegraphics[width=1.0\linewidth]{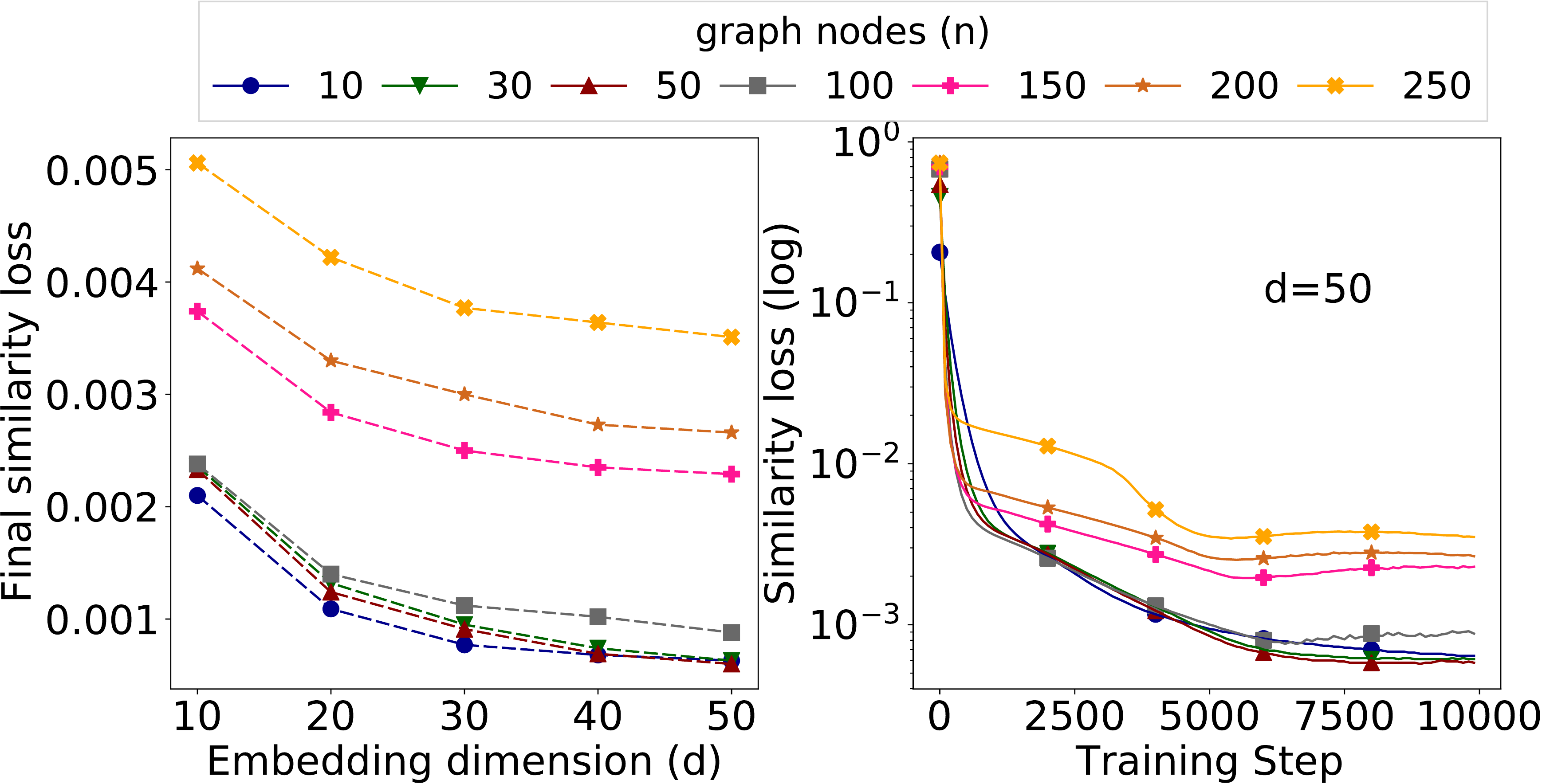}
\caption{Kernel-guided encoder training under different number of nodes and embedding dimension.} 
\label{fig:graph_training}
\end{figure}

\paragraph{Impact of embedding dimension and number of nodes.} Here we conduct an analysis on two factors of the graph encoder: number of nodes $n$ and embedding dimension $d$. We choose $n$ in [10, 30, 50, 100, 150, 200, 250], $d$ in [10, 20, 30, 40, 50], the encoder and decoder are trained for 10k iterations using 50k generated graph pairs. As shown in Figure~\ref{fig:graph_training}, the similarity loss falls fast until convergence. Meanwhile, larger $n$ and smaller $d$ will introduce higher loss. It reveals that our encoder can be applied to search different sizes of graphs by adjusting embedding dimensions.

\noindent\textbf{Impact of proportion of training data.} 
In Table ~\ref{table:ken_tau} and ~\ref{table:pearson}, we show the effectiveness of the proposed embedding scheme by evaluating the impact of different embedding methods with Kendall's tau~\cite{sen1968estimates} and 
Pearson correlation coefficient~\cite{benesty2009pearson}.

We build NASGEM-Bench containing 1000 neural architectures to evaluate the effectiveness of our graph embedding method. 
To minimize the variance from the choice of node operation, each cell in NASGEM-Bench has at most 6 nodes and each node has only one kind of operation, 3$\times$3 convolution. We randomly sample 1000 graphs to build a 3 stage DNN. We train each network on CIFAR-10 with 30 epochs and record the final accuracy.

We use 60\% data (600 graphs) from NASGEM-Bench as the training samples and the rest 40\% (400 graphs) as the test set. Under the 60\% of training data, we randomly slice different proportions of training data for comparison. 
We try three methods to train our predictor based on different proportions of training data. In the first method, we directly flatten the upper triangle of the adjacency matrices, and use the flatten vector and the corresponding performance scores to train the predictor. In the second method, we first use a two-layer MLP to build an autoencoder, then use MLP encoder to embed the adjacency matrices into vectors. We use those vectors and their performance scores to train a predictor. In the third method, we use our kernel-guided encoder to embed the adjacency matrices into vectors, then train the predictor with the scores and embedding vectors. 
Compared with purely using an adjacency matrix and MLP-based autorencoder, NASGEM's embedding method consistently achieves better prediction under different proportion of training data. 

\begin{table}[h!]
\centering
\caption{The Kendall's Tau of the predicted result with different embedding methods.}
\begin{tabular}{lllllllll}
\hline
\multirow{2}{5em}{\shortstack[l]{\textbf{Embedding}\\ \textbf{Method}}}
 & \multicolumn{6}{c}{\textbf{Proportion of Training Data (\%)}}  \\
\cline{2-7}
 & 10      & 20          & 30          & 50          & 70          & 100         \\ \hline
adj. matrix & 0.42                     & 0.45        & 0.44        & 0.44        & 0.46        & 0.46        \\
MLP              & 0.45                   & 0.47          & 0.44          & 0.45          & 0.46          & 0.47          \\ \hline
NASGEM  & \textbf{0.48} &   \textbf{0.53} & \textbf{0.53} & \textbf{0.55} & \textbf{0.54} & \textbf{0.55} \\ \hline
\end{tabular}
\label{table:ken_tau}
\end{table}

\begin{table}[h!]
\centering
\caption{The Pearson correlation coefficient of the predicted result with different embedding methods. }
\begin{tabular}{lllllllll}
\hline
\multirow{2}{5em}{\shortstack[l]{\textbf{Embedding}\\ \textbf{Method}}}
 & \multicolumn{6}{c}{\textbf{Proportion of Training Data (\%)}}  \\
\cline{2-7}
 & 10      & 20          & 30          & 50          & 70          & 100         \\ \hline
adj. matrix & 0.47                     & 0.64        & 0.62        & 0.66        & 0.67        & 0.66        \\
MLP              & 0.57                   & 0.58          & 0.66          & 0.63          & 0.62          & 0.71          \\ \hline
NASGEM  & \textbf{0.76} &   \textbf{0.82} & \textbf{0.84} & \textbf{0.85} & \textbf{0.85} & \textbf{0.85} \\ \hline
\end{tabular}
\label{table:pearson}
\end{table}

\section{Conclusion} 
\workname is the first of its kind estimator-based NAS method that 
tackles down the limitations of graph topology exploration in existing search methods via: (i) construct a topologically meaningful representation by WL kernel guided \textit{graph embedding}; (ii) employ an \textit{efficiency score predictor} to precisely model the relationship between neural architectures and performance; (iii) use \textit{bootstrap optimization} to explore the optimal architecture. 
All these components work coherently to enable \workname to search a more efficient neural architecture from an unrestricted wide search space within a short time.  
Compared with neural architectures produced by existing embedding methods, \netname crafted by \workname consistently achieves higher accuracy on image classification and object detection while having less parameters and MACs. \workname is highly adaptable to different search spaces. 
By combining our proposed graph embedding with NASBench-101, it achieves a more precise and stable prediction compared to the version without graph embedding. 

\bibliography{references}
\bibstyle{aaai}

\end{document}